\documentclass[conference]{IEEEtran}
\IEEEoverridecommandlockouts
\usepackage{cite}
\usepackage{amsmath,amssymb,amsfonts}
\usepackage{threeparttable}
\usepackage{algorithmic}
\usepackage{graphicx}
\usepackage{subfigure}
\usepackage{textcomp}
\usepackage{xcolor}
\def\BibTeX{{\rm B\kern-.05em{\sc i\kern-.025em b}\kern-.08em
    T\kern-.1667em\lower.7ex\hbox{E}\kern-.125emX}}
\begin{document}

\title{Human Semantic Segmentation using Millimeter-Wave Radar Sparse Point Clouds\\
}

% \author{
% \IEEEauthorblockN{
%     Pengfei Song$^{1,3}$\authornote{*}, 
%     Luoyu MEI$^{1,2}$\authornote{*}\thanks{*: Corresponding author}, 
%     Han Cheng$^1$} 
% \IEEEauthorblockA{$^1$ Southeast University, Nanjing, China}
% \IEEEauthorblockA{$^2$ City University of Hong Kong, Hong Kong, China} 
% \IEEEauthorblockA{$^3$ University of California, Berkeley, California, USA\\ Email: [spf, lymei-, 213191775]@seu.edu.cn}
%   } 

\author{
\IEEEauthorblockN{
    Pengfei Song\IEEEauthorrefmark{1}\IEEEauthorrefmark{4}\IEEEauthorrefmark{2}, 
    Luoyu MEI\IEEEauthorrefmark{1}\IEEEauthorrefmark{3}\IEEEauthorrefmark{2}\thanks{\IEEEauthorrefmark{2}: Corresponding author}, 
    Han Cheng\IEEEauthorrefmark{1}} 
\IEEEauthorblockA{
\IEEEauthorrefmark{1} Southeast University,
\IEEEauthorrefmark{3} City University of Hong Kong,
\IEEEauthorrefmark{4} UC Berkeley\\ Email: [spf, lymei-, 213191775]@seu.edu.cn}
  }

\maketitle

\begin{abstract}
This paper presents a framework for semantic segmentation on sparse sequential point clouds of millimeter-wave radar. Compared with cameras and lidars, millimeter-wave radars have the advantage of not revealing privacy, having a strong anti-interference ability, and having long detection distance. The sparsity and capturing temporal-topological features of mmWave data is still a problem. However, the issue of capturing the temporal-topological coupling features under the human semantic segmentation task prevents previous advanced segmentation methods (e.g  PointNet, PointCNN, Point Transformer) from being well
utilized in practical scenarios. To address the challenge caused by the sparsity and temporal-topological feature of the data, we (i) introduce graph structure and topological features to the point cloud, (ii) propose a semantic segmentation framework including a global feature-extracting module and a sequential feature-extracting module. In addition, we design an efficient and more fitting loss function for a better training process and segmentation results based on graph clustering. Experimentally, we deploy representative semantic segmentation algorithms (Transformer, GCNN, etc.) on a custom dataset. Experimental results indicate that
our model achieves mean accuracy on the custom dataset by $\mathbf{82.31}\%$ and outperforms the state-of-the-art algorithms. Moreover, to validate the model's robustness, we deploy our model on the well-known S3DIS dataset. On the S3DIS dataset, our model achieves mean accuracy by $\mathbf{92.6}\%$, outperforming baseline algorithms.

\end{abstract}

\section{Introduction}
Millimeter-wave radar has broad application prospects in the fields of unmanned driving,
 security monitoring, and robotics. Commercial millimeter-wave radars have been deployed.
The semantic information of millimeter wave radar point 
cloud imaging has become a new requirement: in application scenarios 
such as surveillance and human-computer interaction, 
compared with cameras and lidars, millimeter-wave radars have the advantage of 
not revealing privacy, having a strong anti-interference ability, 
and having a long detection distance.

The vast majority of advanced segmentation models (e.g  PointNet\cite{ref1}, PointCNN\cite{ref13}) simply assume that the temporal feature of data is independent of the topological feature. This fails to capture the temporal-topological coupling features under the human semantic segmentation task. We propose a segmentation system that is able to capture the temporal-topological coupling feature. In recent years, significant
effort has gone towards various ways of human semantic segmentation tasks. However, since the data structure of the point cloud loses the topological
feature, such methods (PointNet\cite{ref1}, Point transformer\cite{ref16}) are hard to get trained to fetch satisfying segmentation results during the training process. Thus we construct a graph by weight adjacency matrix to help reconstruct the topological feature when training. In addition, to our best knowledge, 
there does not exist satisfying public datasets and annotations.

Our segmentation system takes advantage of the point cloud processing ability of the neural network. It is able to analyze the characteristics of point cloud information from different angles and arrangements, 
and perform human semantic segmentation on millimeter wave radar point cloud imaging. 
To summarize, our work makes the following contributions:
\begin{itemize}
    \item We propose a framework for semantic segmentation on sparse sequential point clouds of millimeter-wave radar. We are the first to use the opportunity to capture the intrinsic temporal-topological coupling feature to help address the human semantic segmentation problem.
    \item We design a segmentation system to address the challenge of the sparsity and temporal-topological dependence of mmWave point cloud data. We construct a graph by weight adjacency matrix to help reconstruct the topological feature when training. In addition, we collect a new mmWave sparse point cloud dataset (SPC). 
    \item We evaluate our system's performance both on SPC (custom dataset) and the well-known S3DIS dataset. Our model achieves mean accuracy on custom dataset by $\mathbf{82.31}\%$ and outperforms the state-of-the-art algorithms. On S3DIS dataset, our model achieves mean accuracy by $\mathbf{92.6}\%$, still outperforms the baseline algorithms. Thus prove that the system functions stably and captures temporal-topological coupling features successfully.
\end{itemize}

\section{Background and Motivation}
In this section, we first introduce the basic concepts of
human semantic segmentation and then explain the design motivation. Finally, we introduce the net architecture of our work.

\subsection{Human Semantic Segmentation}
Since the input data are points, we need to
process input data to achieve the segmentation labels(i.e., to divide
point clouds into different equivalent classes), consider two
topological space $\mathcal{X}$ and $\mathcal{X}/_\sim$ (Borrowed from the notation
of topology \cite{ref19}, $\mathcal{X}$ is a set of points  $x_i\in\mathbb{R}^n$ with discrete
topology, and $\mathcal{X}/_\sim$ is the set of equivalent classes with the quotient
topology. $\sim$ is the equivalence relation.). And $f^*:\mathcal{X}\to \mathcal{X}/_\sim$ is the ideal function we need to approximate with the help of our model. Thus, it is important to determine the form of equivalent relation $\sim$, i.e., the condition that two points have the same semantic label.

Finally, detecting and analyzing sparse point clouds can be mathematically understood as a function defined on the point cloud space $\mathcal{X}$:
$$\vspace{-0.2em}f^{*}\left(x_{1}, \cdots, x_{n}\right):\mathcal{X}\to\mathcal{X}/_\sim\vspace{-0.2em}$$
where $x_{1}, \cdots, x_{n}$ belong to the point cloud space $\mathcal{X}$. In essence, the task is to find a more suitable neural network to approximate the objective function more accurately $f^{*}$, to achieve a better segmentation effect.
\begin{figure}
	\setlength{\abovecaptionskip}{0.cm}
\centering\subfigure{\label{Fig.sub.1}\includegraphics[width=4cm,height = 5cm]{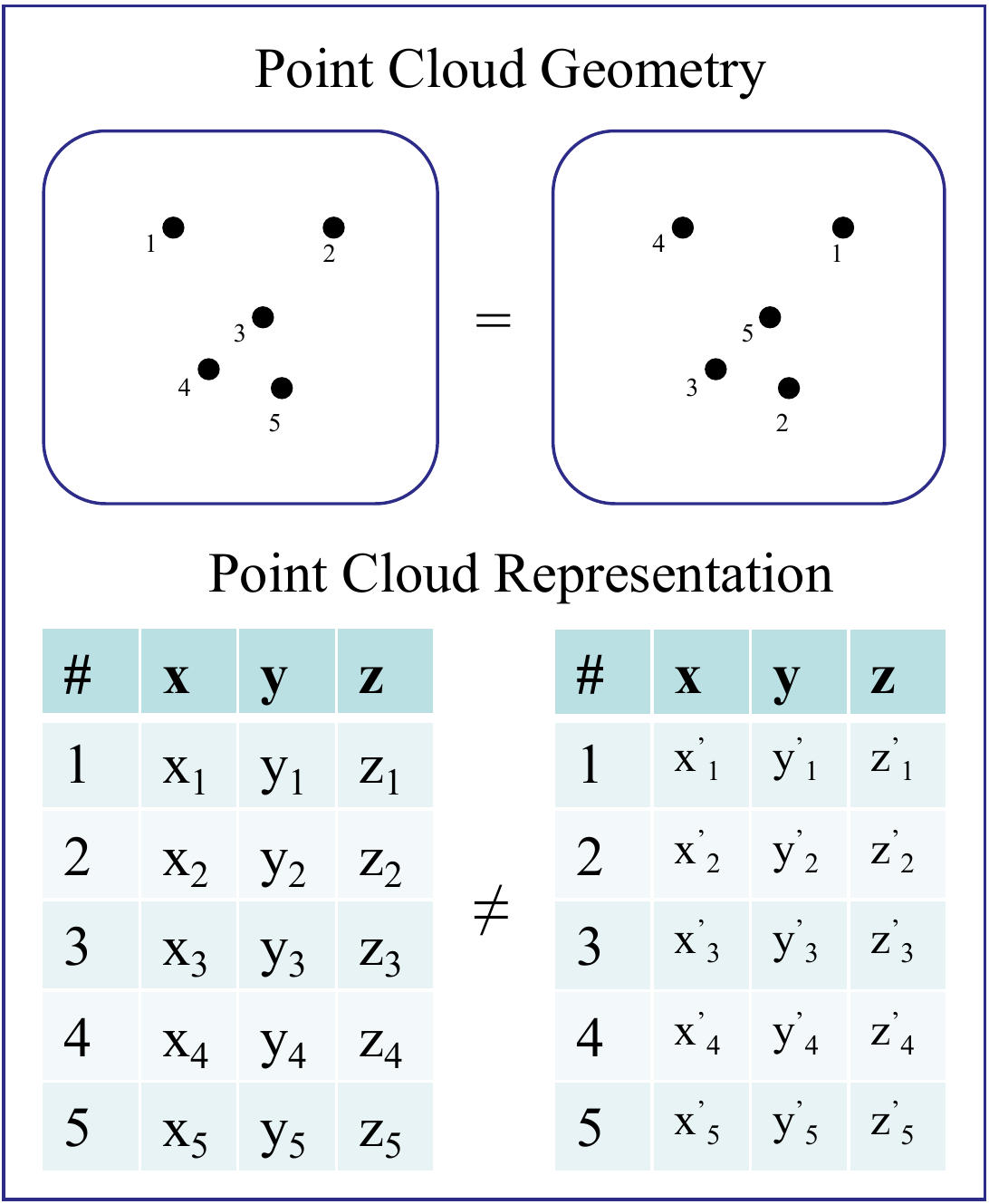}}
\subfigure{\label{Fig.sub.2}\includegraphics[width=4cm,height = 5cm]{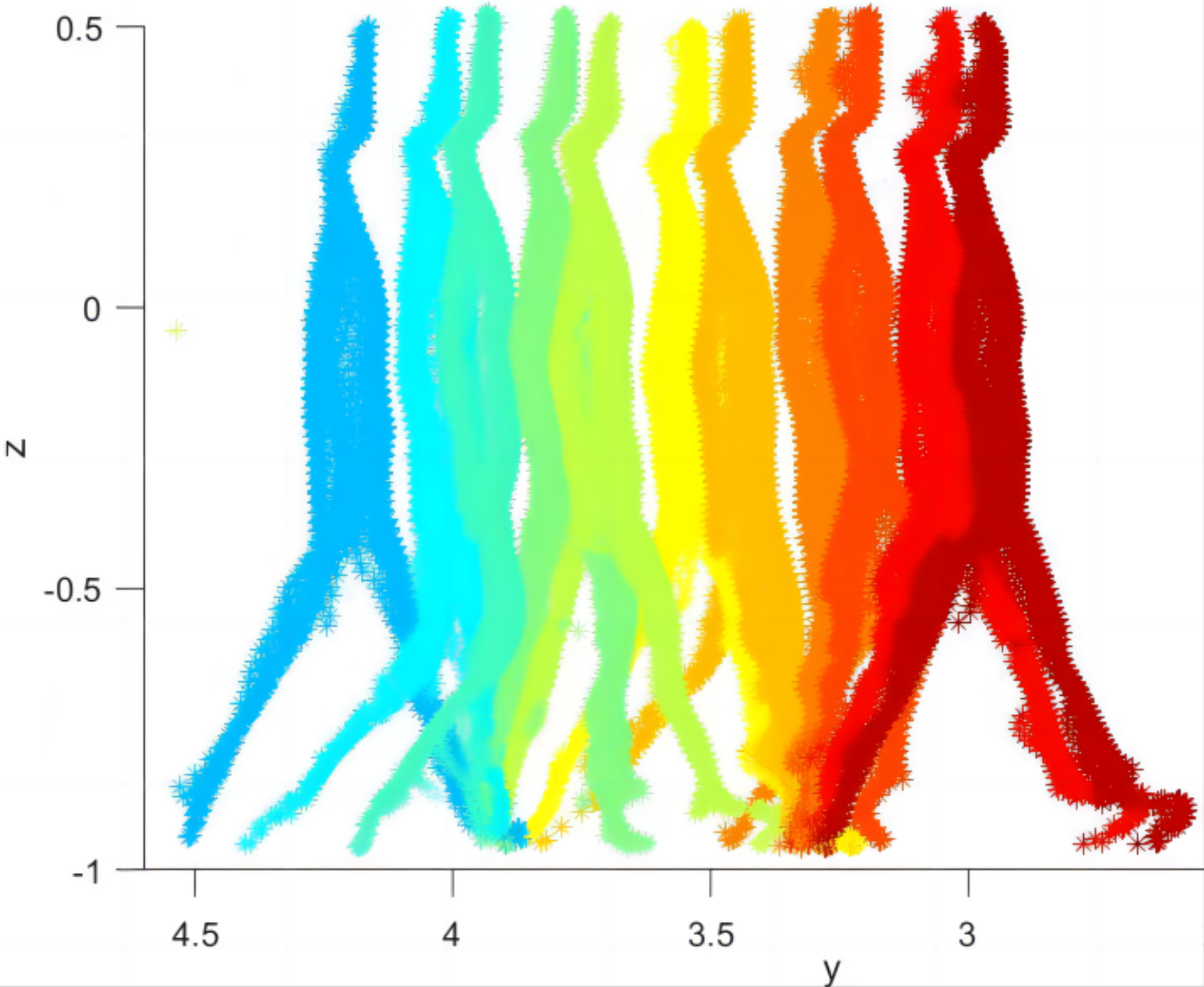}}
	\vspace{-0em}
	\setlength{\belowcaptionskip}{0.cm}
	\caption{Data Structure of Point Cloud. On the left-hand side of Fig.1, we demonstrate the invariance of the order of input point clouds. Detailed description is in section II, part B. On the right-hand side of Fig.1, we illustrate the temporal-topological coupling feature of mmWave human body data. This is a side view of point cloud collected by millimeter-wave radar for the same human body moving forward process. The different colors of the point clouds indicate that the human point clouds belong to different frames. Detailed analysis is in section II, part B.}
	\label{1}
	\vspace{-1em}
\end{figure}

\subsection{Properties of Sparse Point Sets in $\mathcal{X}$}

\begin{figure*}[htbp]
	\includegraphics[width=0.7\textwidth]{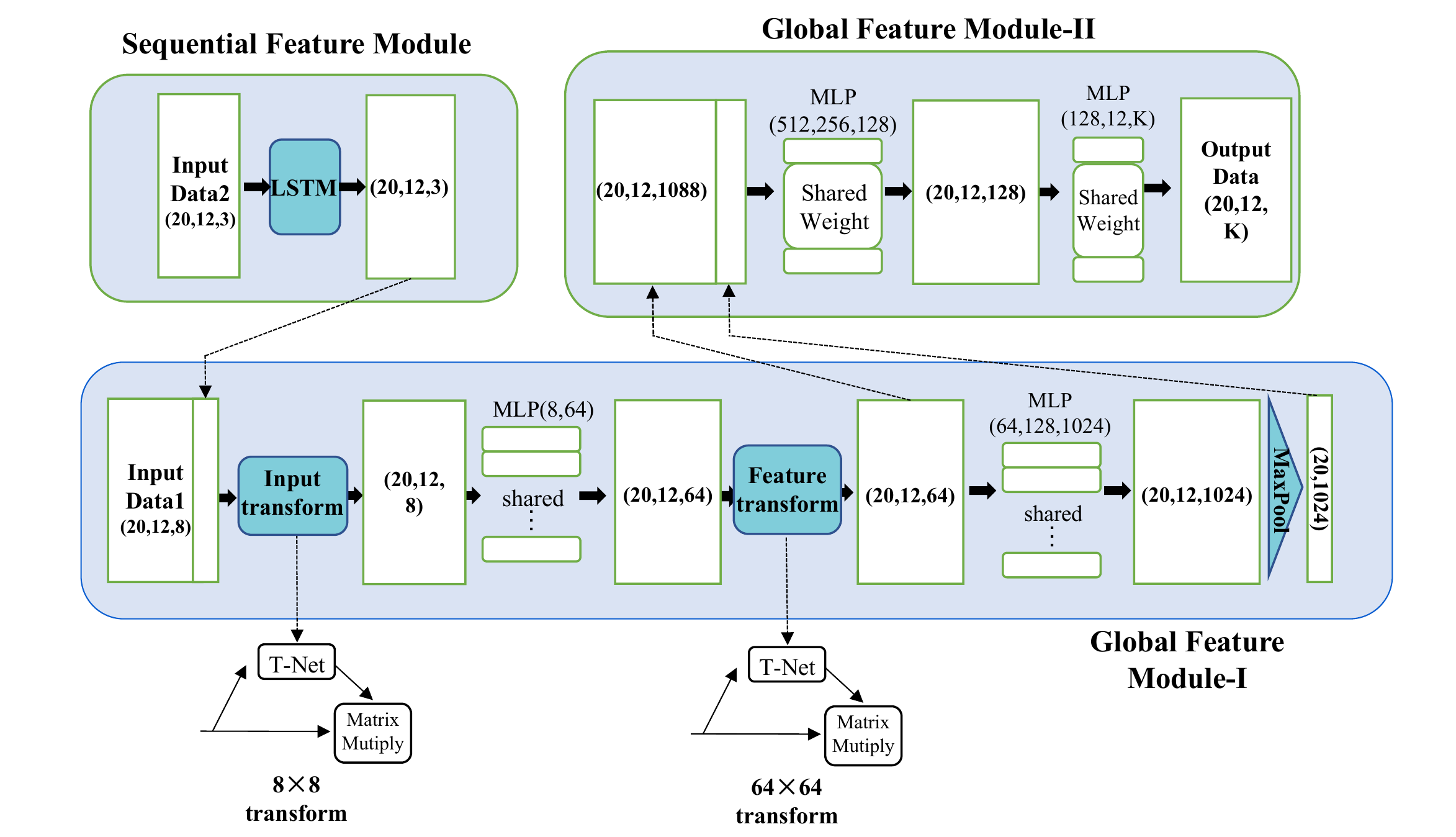}
	\centering
	\caption{Net Architecture of Our Model. Our model contains three parts:   Sequential feature module, global feature module-I, and global feature module II. The number K in Global Feature Module-II is the number of kinds of segmentation labels. Here MLP means multi-layer perceptron. T-NET stands for PointNet\cite{ref1}. The section Sequential Feature Extracting Module further discussed the sequential feature module and analyzed the sequential feature extracting mechanism. The section Global Feature Extracting Module further discussed the global feature module-I.}
	\label{2}
	\vspace{-1em}
\end{figure*}

\begin{itemize}
	\item 
	The sparsity of point clouds: Due to the short wavelength and other reasons, the data collected by millimeter-wave radar is relatively sparse compared with the data collected by traditional lidar.
 %Due to the sparsity, sparse point clouds fairly need more ability to describe the whole picture of an object than dense point clouds.
 %Mathematically, it is reflected in the large variance of the data statistics.
	\item 
	Invariance of the order of input point clouds: As shown on the left-hand side of Fig.1, unlike other traditional methods of image processing and so on, the result of $f^{*}\left(x_{1}, \cdots, x_{n}\right)$ is invariant to certain transformations (e.g., permutation) of input data. Given a set of points, what point cloud geometry means that the serial number doesn’t matter to the geometry of points. Whatever the serial number of a point is, the point is still itself geometrically. But if we change the serial number of each point, when we represent the point clouds by matrix, the point cloud representation will change.
	\item 
	Temporal-topological coupling feature: As shown on the right-hand side of Fig.1, a bunch of frames of the human body moving forward process data are included. Different frames are distinguished by different colors. Due to the discontinuity of the data collecting process, one part of the human body is not continuously tracked. Take feet for example, it is not fully captured in all different frames of the same human. Considering that the point clouds contain the information of coherent actions like walking or arm-swinging, the data is temporally-topologically related:
		$$
		f\left(\pi\left\{x_{1}^{(k)}, \ldots,x_{m}^{(k)}\right\}\right) \approx f\left(\pi\left\{x_{1}^{(k+1)}, \ldots, x_{m}^{(k+1)}\right\}\right)
		\vspace{-0.2cm}$$
	where $x_{m}^{(k)}$ stands for a point, m means the m-th point in the point cloud of one frame, and k means the k-th frame in the time series. The feature-extraction projection  $\pi$  maps point clouds from Euclidean space to time series feature spaces.
\end{itemize}
\subsection{Net Architecture}
There are three key parts in our network: the sequential feature module, the global feature module-I for points' global feature extraction, and the global feature module-II for semantic segmentation results' process as well as output.
\begin{itemize}
	\item 
	The architecture of our network is demonstrated in Fig.2, which contains targeted effective solutions for sequential feature extraction. A more detailed introduction of our network(parameters and so on) is included in the caption of Fig.2.
	\item 
	Considering that point cloud sequences %$\vspace{-0.3cm}$$$\vspace{-0.3cm}\{p_1,p_2,\dots,p_k,\dots\}$$ (where $p_k=\{x_1^{(k)},\dots,x_m^{(k)}\}$,) 
 have sequence dependencies, the semantic information of different point clouds interferes with each other. To solve the problem, we proposed targeted solutions in the sequential feature module in Fig.2.
\end{itemize}

\section{Main Design}
In this section, we first describe capturing temporal-topological features of mmWave data, then
explain the global feature extracting module of our model. Then, we explain the construction of custom loss function by graph theory to help reconstruct the topological feature when training. Finally, we introduce the multitask method used when training.

%%\subsection{Theoretical Analysis}
%\subsubsectsequentiallyty of point clouds}
%Assume there are two different point clouds of the same object: $\mathcal{D}$ and $\mathcal{S}$. $\mathcal{D}$ stands for dense point clouds , while $\mathcal{S}$ repesents sparse point clouds. Consider two ideal segmentation functions $\mathcal{F}$ and $f$. $\mathcal{F}$ consumes dense data from  $\mathcal{D}$ and $f$ consumes sparse data from  $\mathcal{S}$. To solve the problem, we need to propose a network that can approximate $f$ theoretically and practically. Besides, we could describe those concepts in mathematical way:
%$$
%\mathcal{F}:\mathcal{D} \rightarrow \mathcal{F}(\mathcal{D})
%$$
%$$
%f:\mathcal{S} \rightarrow f(\mathcal{S})
%$$
%if
%$$
%\mathcal{S} \subset \mathcal{D}
%$$
%then
%$$
%f(\mathcal{S}) \subset \mathcal{F}(\mathcal{D})
%$$
%where$\mathcal{F}(\mathcal{D})$ means the segmentaion results of points clouds $\mathcal{D}$, and $f(\mathcal{S})$ means the segmentation results of point clouds $\mathcal{S}$. Noticing that $\mathcal{D}$, $\mathcal{S}$ , $\mathcal{F}(\mathcal{D})$ , $f(\mathcal{S})$ are all sets.
%\subsubsection{Sequential semantic information relevance of point clouds}
%Considering that our data collects a coherent sequence of motions, we could describe our data in a mathematical way:
%$$\left\{x_{1}^{(k)}, \ldots, x_{m}^{(k)}\right\}$$
%where k is the sequence label. Noticing that the points set belongs to space $R^{m \times n}$, where m means the number of points and n means the feature dims of a point.

\subsection{Sequential Feature Extracting Module} 
Inspired by \cite{wang-tkde-parking}, we illustrate our sequential feature-extracting mechanism in Fig.3 and describe the formulation as below:
\vspace{-0.4cm}
\begin{equation}
    \vspace{-0.3cm}f_t=\sigma(W_f\cdot[h_{t-1},x_t]+b_f)\!
\end{equation}
\begin{equation}
\vspace{-0.3cm}i_t=\sigma(W_i\cdot[h_{t-1},x_t]+b_i)\!
\end{equation}
\begin{equation}
    \vspace{-0.3cm}\Tilde{C_t}=tanh(W_c\cdot[h_{t-1},x_t]+b_c)\!
\end{equation}
\begin{equation}
    \vspace{-0.3cm}C_t=f_t*C_{t-1}+i_t*\Tilde{C_t}\!
\end{equation}
\begin{equation}
    \vspace{-0.3cm}O_t=\sigma(W_o\cdot[h_{t-1},x_t]+b_o)\!
\end{equation}
\begin{equation}
    \vspace{-0.3cm}h_t=O_t*tanh(C_t)\!
\end{equation}
The point cloud data is sparse as well as sequential.
%which means that $\exists\epsilon\in N$, s.t $||x_t-x_{t-1}||>\epsilon$. ($N$ stands for positive real number set) 
Since the mechanism has Eq.(4), which is able to hold long-term memories, thus solving the sequential-related problem. 

  \begin{table}[!ht]
  \vspace{-1em}
    \caption{Math Notitions}
    
    \centering
    \setlength{\tabcolsep}{7mm}{
    \begin{threeparttable}          %这行要添加
      \begin{tabular}{cc}
				\hline
				Notition&Description\\
				\hline
				$\sigma$ &Activation Function\\
                    $W_{\cdot}\tnote{1}$ &Trainable Weights\\
                    $b_{\cdot}\tnote{2}$ &Bias\\
                    $C_t$ &Long-Term Memory\\
                    $f_t$ &Forgetable Weight\\ 
                    $O_t$ &Output with Sequential Feature\\
                    $h_t$ &Final Output\\
                \hline
      \end{tabular}
        \begin{tablenotes}
	    \footnotesize
	    \item[1] $W_{\cdot}$ is generalized description of $W_f$, $W_i$, $W_c$. 
            \item[2] $b_{\cdot}$ is genralized description of $b_f$, $b_i$, $b_c$.
        \end{tablenotes}
    \end{threeparttable}
    }
    \label{Tab1}
    \vspace{-1em}
  \end{table}

\begin{figure}
	\setlength{\abovecaptionskip}{0.cm}
	\centering
	\includegraphics[width=6cm,height=4cm]{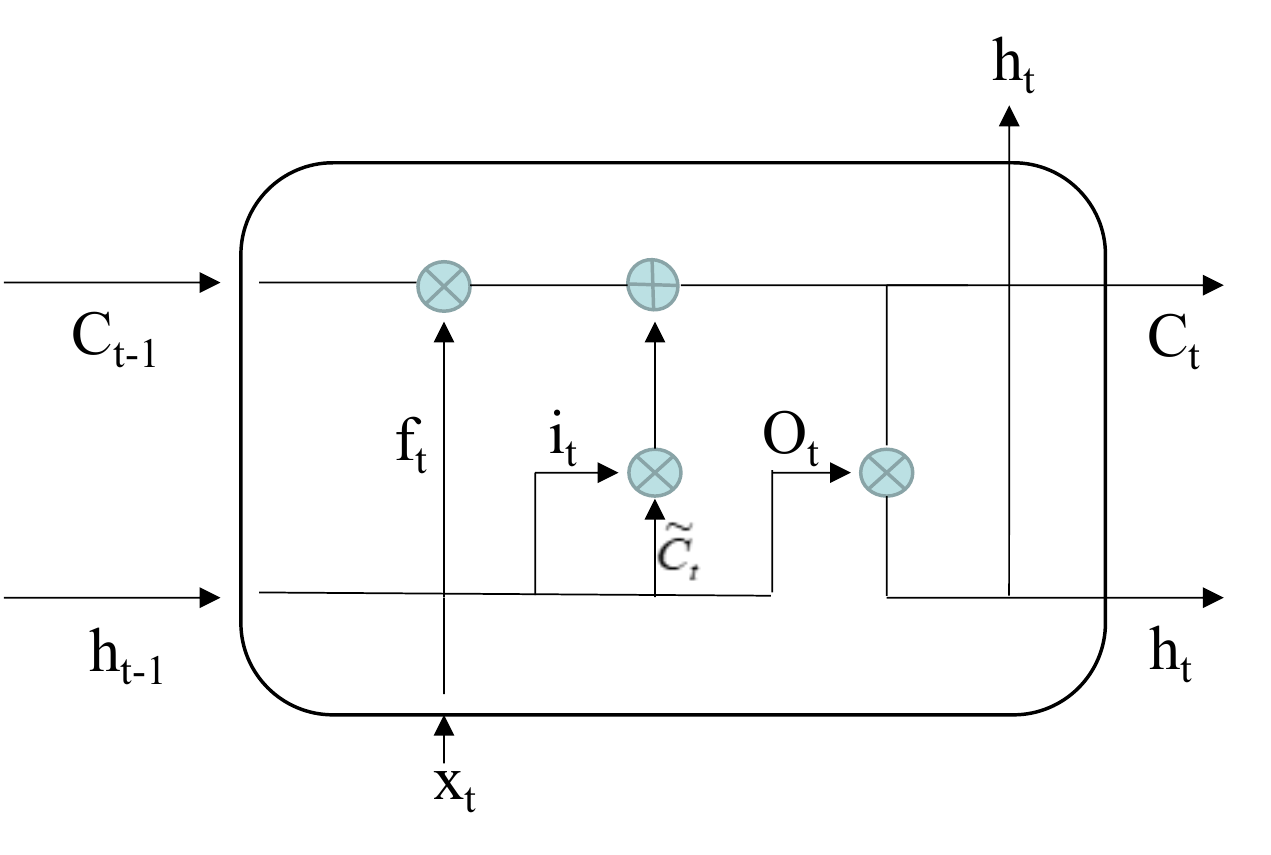}
	\vspace{-0.5em}
	\setlength{\belowcaptionskip}{0.cm}
	\caption{Sequential Feature Extracting Mechanism. More detailed math notitions are included in Table \ref{Tab1}}
	\label{3}
        \vspace{-1em}
\end{figure}
$\vspace{-0.5cm}$
\subsection{Global Feature Extracting Module}
%Overfitting is overviewed as phenomenon when\\$\vspace{-0.5em}$
%\begin{small}
  %  $$\vspace{-1em}\frac{1}{2N}\sum_{i=1}^{N}(h_{\theta}(x_{train}^{(i)})-y_{train}^{(i)})^2\searrow 0$$
%\end{small}we get\\$\vspace{-1em}$
%\begin{small}
 %   $$\frac{1}{2N}\sum_{i=1}^{N}(h_{\theta}(x_{test}^{(i)})-y_{test}^{(i)})^2\nearrow$$
%\end{small}

%\\(where h is our model, $\theta$ are the model parameters, $x_{train}$ and $y_{train}$ are the training point cloud data and training point cloud data label, respectively.)
%\\Thus defining the bias function: 
%$\vspace{-0.3cm}$
%\begin{equation}
   % \vspace{-0.4cm}\sigma(\theta,x,y)=\frac{1}{2N}\sum_{i=1}^{N}(h_{\theta}(x^{(i)})-y^{(i)})^2
%\end{equation}
%\\Next, when input data is dense, we want to show that overfitting occurs more often.
In Global Feature Module I and Module II, we compute a global signature $F\in \mathbb{R}^K$ (K is the global feature dimension) for points according to each frame of the human body data sequence. Then we feed the global signature $F$ back to per point features by concatenating the global signature $F$ with each of the point features. 
%And the ability of global feature extracting of our model goes deeper as well as stronger at the same time. 
Then we repeat the action to make our model extract global features deeply enough. Given a point with its global signature, since each point belongs to its human body frame, we could make better semantic segmentation results if we have more wise global features for this particular point. Note that we should master the depth of the global extracting module since overfitting will occur more often when the neural network goes deeper.

\subsection{Design of Custom Loss Function}
%Since the data structure of the point cloud loses the topological feature, we construct a graph by weight adjacency matrix to help deal with this problem. In conclusion, 
The segmentation task requires us to find the partition of the graph such that the weight between different groups of vertices on the graph is quite low as well as similar to the target segmentation results. Since the points with the same label should have rather a strong relationship with each other, the weights within the same kind of points should have high weights. As illustrated in Fig.4, suppose the segmentation output (shown in the middle graph of Fig.4) is that points 1, 2, and 5 are of the same kind. Point 3 and point 4 are of another 2 different kinds. Besides, suppose the target segmentation result (shown on the right of Fig.4) is that points 1 and 2 are of the same kind. Points 3 and 5 are of the same kind. Point 4 is of the other different kind. Thus a graph is partitioned into different groups, and each group of points has the same label.
\\How to describe the loss? The original idea is to weigh the similarity of the output and target result. Thus we define the connectivity of a partitioned graph:$\vspace{-0.3cm}$
\begin{equation}
    \vspace{-0.3cm}C = \sum_{i=1}^{n}\sum_{j=i}^{n}\delta_{ij}W_{ij}
\end{equation}
Here $n$ is the number of points in the point cloud,
\\ $\delta_{ij}=
\begin{cases}
	0& when\; point\; i\; and\; j\; are\; of\; the\; same\; kind\\
	1& when\; point\; i\; and\; j\; are\; of\; different\; kind
\end{cases}$ 
\\and $\delta$ function is determined by the segmentation label, which means that given two same graphs with different labels like the middle and the right graph in Fig.4, though sharing the same weight matrix, their coordinating $\delta$ function will be different.
\\And $W$ stands for the weight adjacency matrix computed according to the Gauss Similarity Function:
\\$W_{ij}=exp^{-{\Vert p_i - p_j \Vert_2}^2}$, where $p_i,p_j\in \mathbb{R}^{d_x}$, $d_x$ is the feature dimensionality of each point, $p_i,p_j$ is the point i and point j in the graph, the norm of $p_i - p_j$ is 2-norm.
\\Finally, the custom self-designed graph loss function is shown below, which introduces topological feature into the point cloud segmentation task:
$\vspace{-0.5cm}$
\begin{equation}
    \vspace{-0.34cm}C_1 = \sum_{i=1}^{n}\sum_{j=i}^{n}\delta^1_{ij}W_{ij}
\end{equation}
\begin{equation}
    \vspace{-0.3cm}C_2 = \sum_{i=1}^{n}\sum_{j=i}^{n}\delta^2_{ij}W_{ij}
\end{equation}
\begin{equation}
    \vspace{-0.6cm}L=a^{|C_1-C_2|}-1
\end{equation} 
\\Here $C_1$ and $C_2$ are computed according to the connectivity formula above, but their $\delta$ function is different. $C_1$ stands for the connectivity of our model's segmentation result, as illustrated in the middle of Fig.4. $C_2$ stands for the connectivity of the target segmentation result, as illustrated in the right of Fig.4. Because the two graphs' segmentation label is different, so their $\delta$ function is different. Empirically, we set $a = 1.1$ in experiments.
\begin{figure}
	\setlength{\abovecaptionskip}{0.cm}
	\centering
	\includegraphics[width=9cm,height=3cm]{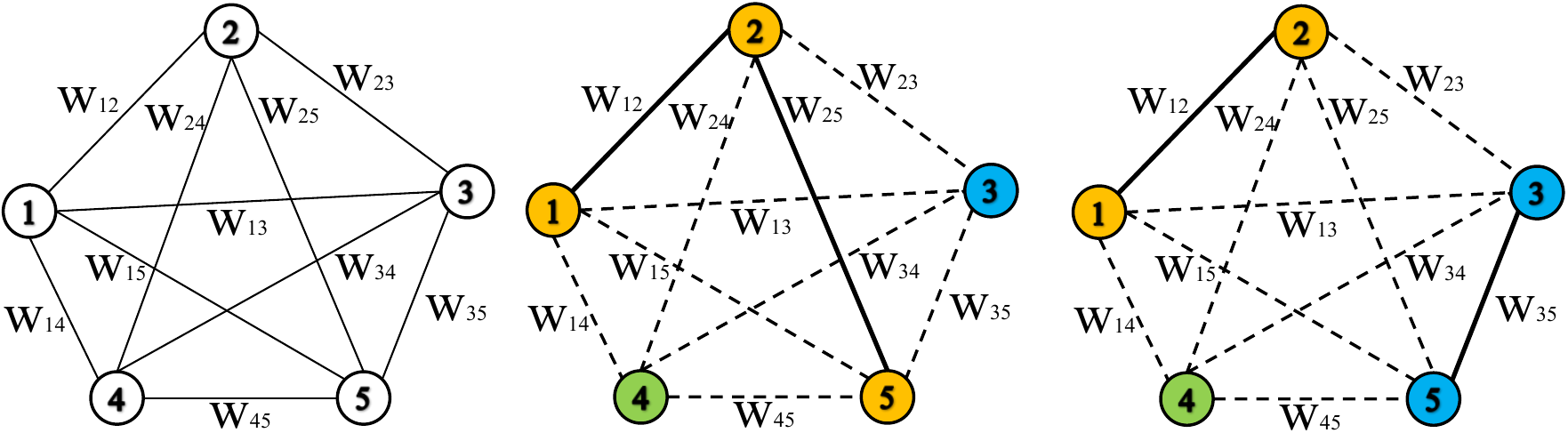}
	\vspace{-1em}
	\setlength{\belowcaptionskip}{0.cm}
	\caption{Segmentation Visualization on Graphs.  
		This is a complete graph of five points. The item sitting on the left is the weight matrix visualization on the graph. $W_{ij}$ means the weight between points i and j. The middle item is the schematic diagram of our model's output. Points with the same label are painted in the same color and connected with solid black lines. Points with different labels are connected with a dotted line. Different colors mark the different labels of points.  The right item is the schematic diagram of the target segmentation result.}
	\label{3}
	\vspace{-1em}
\end{figure}
\subsection{Training with Multitask}
The training data is sparse and sequence-dependent. We labeled the point clouds as head, prothorax, left arm, right arm, left leg, and right leg. Some semantic labels have similar semantic information, like left leg and right leg, which means that classifying those semantic-similar points requires more effort. To improve the effectiveness of the six classifications, we also labeled the point clouds as head+prothorax, arm, and leg to do Multi-task learning. 
\section{Experiments}

\begin{table*}[htbp]

\caption{Semantic Segmentation Results on mmWave Sparse Point Cloud Dataset}
\vspace{-1.5em}
\begin{center}
\begin{threeparttable}
\begin{tabular}{|c|c|c|c|c|c|c|c|c|}
\hline
Network&mAcc&mIoU&Head&Chest&Left Arm&Right Arm&Left Leg&Right Leg\\
\hline
DGCNN&81.67&65.58&89.34&77.99&82.69&79.78&80.72&78.81\\
\hline
PointNet&80.96&64.82&91.32&80.80&79.78&80.72&83.05&71.67\\
\hline
mmMesh&79.80&61.72&90.74&79.18&78.45&78.75&78.21&74.99\\
\hline
PointCNN&77.27&45.67&93.77&47.81&86.18&80.36&80.90&72.91\\
\hline
SetTransformer&62.67&54.15&80.34&100.00&59.83&55.68&44.43&33.86\\
\hline
Ours&82.31&67.45&91.18&81.86&81.09&77.39&80.00&78.55\\
\hline
\end{tabular}

\label{Tab2}
\begin{tablenotes}
	\footnotesize
	\item Metric is mIoU and mAcc(mean accuracy) on points. We compare with outstanding traditional methods. Our method achieved state-of-the-art in both mIoU and mAcc. Detailed analysis is in part IV, section B.
\end{tablenotes}

\end{threeparttable}
\end{center}
\end{table*}

\begin{table*}[htbp]
\vspace{-2.5em}
	\caption{Semantic Segmentation Results on S3DIS}
        \vspace{-2.5em}
	\begin{center}
		\resizebox{\textwidth}{!}{
			\begin{tabular}{|c|c|c|c|c|c|c|c|c|c|c|c|c|c|c|c|c|}
				\hline
				Network&mAcc&mIoU&Ceiling&Floor&Wall&Beam&Column&Window&Door&Table&Chair&Sofa&Bookcase&Board&Clutter\\
				\hline
				MinkowskiNet&71.7&65.4&91.8&98.7&86.2&0.0&34.1&48.9&62.4&81.6&89.8&47.2&74.9&74.4&58.6\\
				\hline
				PointNet&49.0&41.1&88.8&97.3&69.8&0.1&3.9&46.3&10.8&59.0&52.6&5.9&40.3&26.4&33.2\\
				\hline
			    KPConv&72.8&67.1&92.8&97.3&82.4&0.0&23.9&58.0&69.0&81.5&91.0&75.4&75.3&66.7&58.9\\
				\hline
				PointCNN&63.9&57.3&92.3&98.2&79.4&0.0&17.6&22.8&62.1&74.4&80.6&31.7&66.7&62.1&56.7\\
				\hline
				PointTransformer&76.5&70.4&94.0&98.5&86.3&0.0&38.0&63.4&74.3&89.1&82.4&74.3&80.2&76.0&59.3\\
				\hline
				Ours&92.6&83.5&75.6&79.9&98.4&98.1&99.8&93.8&98.0&98.1&99.8&96.0&99.5&89.8&74.2\\
				\hline
			\end{tabular}
		}
		\label{Tab3}
	\begin{tablenotes}
	\footnotesize
	\item Metric is mIoU and mAcc (mean accuracy). Our method achieved the state-of-the-art in mIoU and mAcc.
\end{tablenotes}
		
	\end{center}
 \vspace{-0.4cm}
\end{table*}

\subsection{Dataset and Data Collection}
\subsubsection{mmWave Sparse Point Cloud Dataset (SPC)}
To our best knowledge, there doesn't exist satisfactory dataset opening to the public. 
We collect the SPC data using mmWave radar. In the data-collecting process, we use the iwr6843 commercial mmWave radar, which emits FMCW signals from
its transmitting antennas and captures the reflected signals using its
receiving antennas. Then the radar hardware mixes the received
signals with the transmitted signal to obtain the IF (Intermediate
Frequency) signals, which are the outputs of the mmWave radar. The data collecting process is shown in Fig.5.
\\In addition to the collection of mmWave data, we also
use the Kinect as the motion-capturing device to obtain high precision
dynamic pose information of the subject, which is utilized to generate the ground truth point cloud that is used to train the
proposed deep learning model in our system. There are human body data categorized into 6 kinds: Head, chest, left arm, right arm, left leg, and right leg. The dataset has over 11,000 frames of human body data, which is split into  70 percent for training, 20 percent for testing and 10 percent for validation.
\begin{figure}
	\setlength{\abovecaptionskip}{0.cm}
	\centering
	\includegraphics[width=7cm,height=4cm]{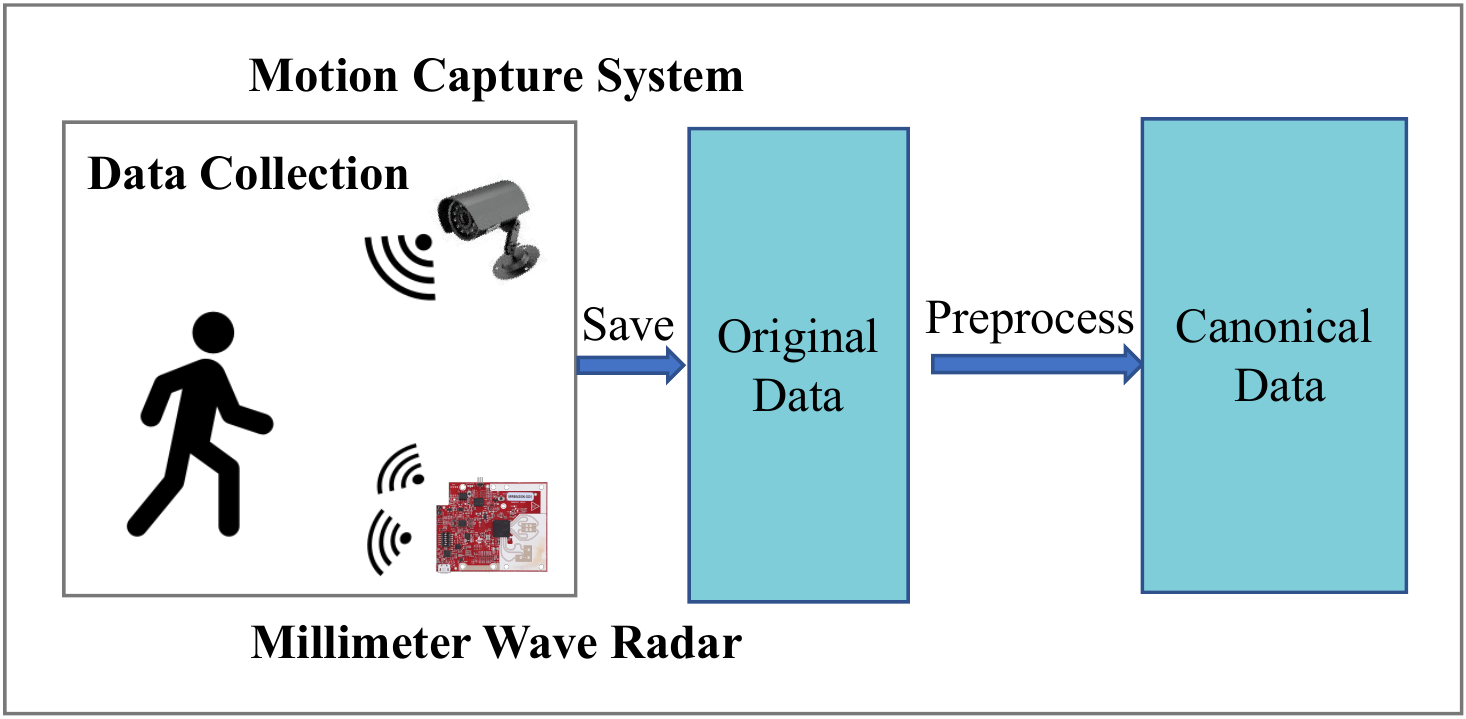}
	\vspace{-0em}
	\setlength{\belowcaptionskip}{0.cm}
	\caption{Original Data Collection and Preprocessing.  
	}
	\label{4}
\end{figure}

\subsubsection{S3DIS Dataset}
To validate the robustness of our model, we deploy our model on the well-known S3DIS dataset.

\subsection{Results on SPC Dataset}
We evaluate our model together with other outstanding models on the mmWave Sparse Point Cloud Dataset. There are human body data categorized into 6 kinds: Head, chest, left arm, right arm, left leg, and right leg. While previous methods are called masterpieces of various branches in the field of point cloud segmentation, to our best knowledge,
we are the first to train with a newly designed graph loss function.

The segmentation results are shown in Table \ref{Tab2}. And the experiments in Table \ref{Tab2} indicate that our model outperforms the baseline algorithms. To address the problem caused by the sparsity of input data, the original idea is to improve the ability of feature extraction. We lift the feature dimension for our model to grab features globally. Under these circumstances, it is worthy stressing that we should focus less on the local signature of each point and pay more attention to feature extracting as well as fixing local feature mistakes. That is determined by the sparsity of input data, which means that compared to extracting global features, over-fitting occurs more often when extracting local features, which is shown in the experiment(in Table \ref{Tab2}). The Transformer model and other models using the self-attention mechanism don't perform well. We analyze the reason is that the attention mechanism focuses on the local feature of each point, thus leading to over-fitting mistakes and reducing the segmentation results. 

\subsection{Results on S3DIS}
Besides, to validate the robustness of our model, we deploy it on the  Stanford 3D semantic parsing dataset \cite{ref20}.
The segmentation results are demonstrated in Table \ref{Tab3}. Our model is tested on the S3DIS dataset and achieves an excellent result, which shows that our model is robust as well as efficient. Compared to previous studies, our method outperforms them
by a large margin on the beam categories reported.
\subsection{Experiments of Custom Loss Function}
To show that our newly designed graph loss function is more fitting with the point cloud segmentation result, we draw the variation of training accuracy using different loss functions w.r.t training epoch in Fig.6. From the curves we say that the accuracy using custom loss function converges faster especially when training epoch increasing from 40 to 60. It is easier for our model to improve performance after introducing graph structure in the loss function.
\begin{figure}
	\setlength{\abovecaptionskip}{0.cm}
	\centering
	\includegraphics[width=8cm,height=6cm]{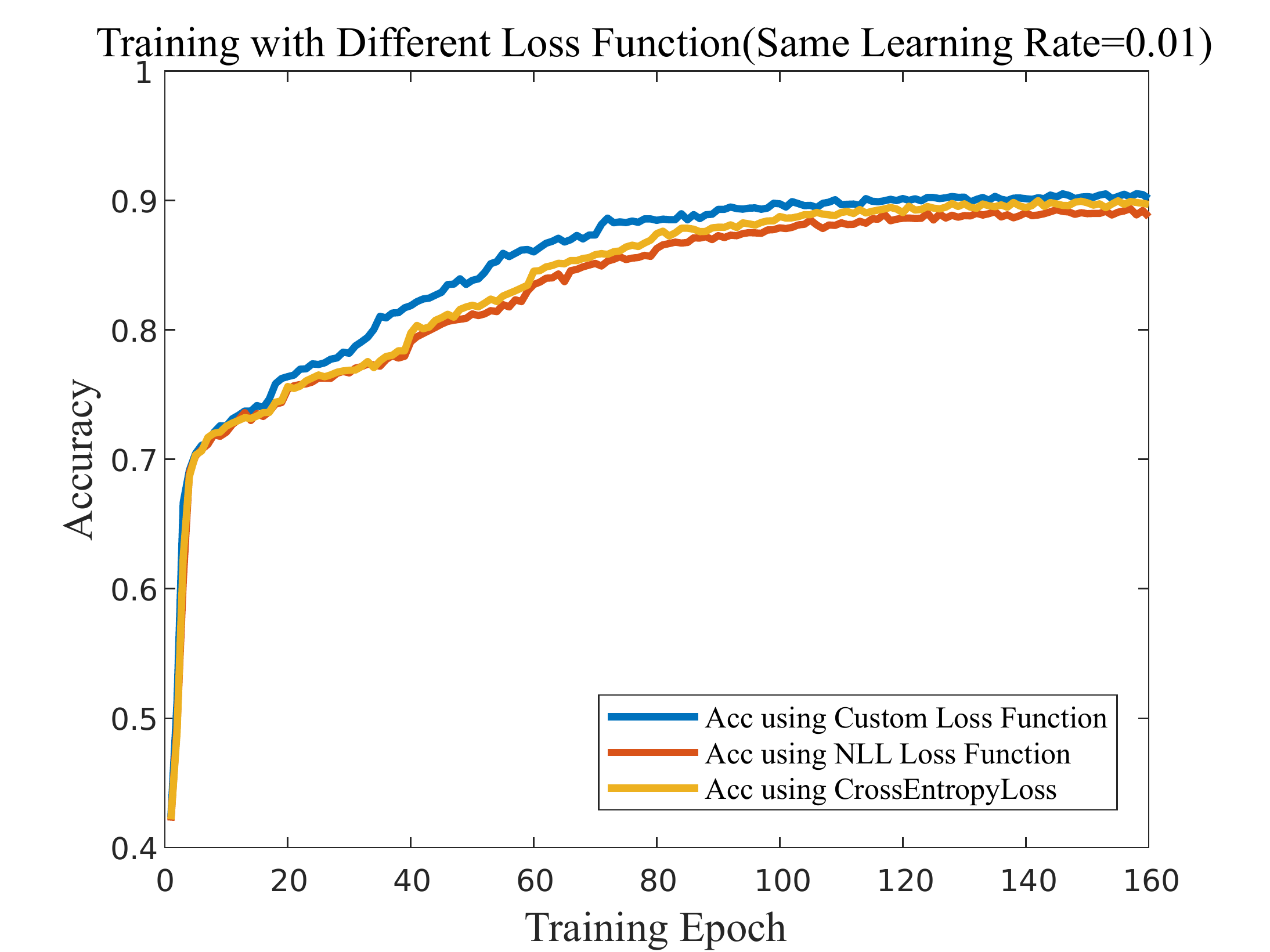}
	\vspace{-0em}
	\setlength{\belowcaptionskip}{0.cm}
	\caption{Custom Train Loss Function.  
	}
	\label{6}
 \vspace{-2em}
\end{figure}

\section{Related Work}
\subsection{Pioneer Work}
Authors at Stanford University propose the PointNets series (including PointNet\cite{ref1} to understand the three-dimensional spatial characteristics of objects, realizes object classification and scene segmentation, which is a pioneering work for directly processing point sets. PointNet\cite{ref1} was first proposed in 2016. Its basic architecture consists of multiple layers of perception.  However, the design of its network structure makes it difficult to capture the local features of the point cloud as well as to process the point cloud sequence data related to time series.
\subsection{Graph-based Network}
Graph neural network-based studies~\cite{yan20222,liu2022pred,liu2022exploiting,ref4,ref5} have developed rapidly in recent years.
In point cloud processing, Kipf et al. \cite{ref4} proposed a graph convolutional network (GCN) for semi-supervised learning on graph structure data, which builds a local graph for each point group in the network layer and aggregates point data according to the relationship between points. Kim et al.\cite{ref5} proposed a rotation-invariant recognition model RI-GCN for 3D point cloud data, based on Graph convolutional neural networks (GCNs). 
% Xu et al. \cite{ref6} proposed Grid-GCN for fast and scalable point cloud learning. Coverage aware Grid query (CAGQ) is also introduced in Xu et al. \cite{ref6}.
\subsection{Continuous Convolution Network}
Lyu et al. \cite{ref7} project a 3d point cloud into 2d image space, enabling traditional 2D convolutional neural networks to be used for segmentation.
However, the results of this method are not satisfactory on some datasets. PointCNN\cite{ref13} arranges point clouds into normative order by introducing  $\mathcal {X} $-transformation, and then performs the element-by-element product sum operation of typical convolution operators on  $\mathcal {X} $-transformed features. Although the effect of  $\mathcal {X}$-transformation is not as expected, it still improves the accuracy of prediction.
\subsection{Attention-based Network}
Attention network \cite{ref15} uses the Attention mechanism for machine translation tasks, and achieved good results. 
%Due to the similarity between the point cloud and semantic sequence in the data structure (both can be expressed in the form of tensors), the attention-based point cloud processing network begins to show a bright future. 
SetTransformer\cite{ref16} and PointTransformer\cite{ref17} are typical representations of the attention-based network. PointTransformer\cite{ref17} designed a self-attention layer for the point cloud and applied it to the semantic segmentation of the point cloud, achieving excellent results. SetTransformer\cite{ref16} designed its encoder and decoder based on the attention mechanism. 
%To reduce computational complexity, the inductive point method is introduced inspired by the sparse Gaussian process in SetTransformer \cite{ref16}.
\subsection{Sparse Segmentation Network}
Compared with the lidar point cloud, the millimeter-wave radar point cloud is very sparse, which makes the point cloud lose some spatial characteristics.
Li et al. \cite{ref8} proposed a fast segmentation algorithm for the sparse point cloud. Schumann et al. \cite{ref9} accumulated multiple radar frames to obtain richer point clouds, avoiding the sparse characteristics of millimeter-wave radar point clouds, and then applied the Frustum PointNet with a small adaptation to segmentation.
\subsection{Other Related Work}
Adhikari et al. \cite{ref11} proposed a full-body contour imaging and three-dimensional pose estimation system named MilliPose. mesh \cite{ref18} used the pointnet-based template to preliminarily process the human point cloud collected by millimeter-wave radar, and then input the preliminarily processed data into the SMPL model to obtain a mesh human model for attitude estimation as well as semantic segmentation.

\section{Conclusion}
In this paper, we propose an efficient and outstanding framework for semantic segmentation on sparse sequential point clouds. Our model is capable of achieving satisfying segmentation results under the situation of sparse, sequential, and non-structured data. Moreover, since the point cloud data lack topological features, we also analyzed and optimized the data structure of the point cloud. To our best knowledge, there are no satisfying mmWave datasets with sparse feature open to the public. To cope with that, we collected SPC (custom dataset) scientifically and cautiously. 
We experimentally show that our model outperforms many existing segmentation models both on the famous S3DIS dataset and the custom dataset. We wish this work to pave the way for future research on human semantic segmentation.
%To make a comparison, we deploy some other outstanding semantic segmentation algorithms (Transformer, GCNN, .etc) on our own dataset. %And the experiments in Table \ref{Tab2} turns out to show that our model outperforms the chosen algorithms. Besides, to validate the robustness of our model, we deploy our model on the well-known S3DIS dataset. The experiment results in Table \ref{Tab3} show that our model is stable and still outperforms the chosen algorithms.

\section{Acknowledgement}
This work was supported in part by 2030 major projects of scientific and technological innovation under Grant No. 2021ZD0114200, National Natural Science Foundation of China under Grant No. 62272098.


\begin{thebibliography}{00}
\bibitem{ref1}Qi, C., Su, H., Mo, K., \& Guibas, L.J. (2017). PointNet: Deep Learning on Point Sets for 3D Classification and Segmentation. IEEE Conference on Computer Vision and Pattern Recognition (CVPR), 77-85.
%\bibitem{ref2}Qi, C., Yi, L., Su, H., \& Guibas, L.J. (2017). PointNet++: Deep Hierarchical Feature Learning on Point Sets in a Metric Space. NIPS.
%\bibitem{ref3} Qi, C., Liu, W., Wu, C., Su, H., \& Guibas, L.J. (2018). Frustum PointNets for 3D Object Detection from RGB-D Data. 2018 IEEE/CVF Conference on Computer Vision and Pattern Recognition, 918-927.
\bibitem{ref4} Kipf, T. N., \& Welling, M. (2016). Semi-supervised classification with graph convolutional networks. arXiv preprint arXiv:1609.02907.
\bibitem{ref5} Kim, S., Park, J., \& Han, B. (2020). Rotation-invariant local-to-global representation learning for 3d point cloud. Advances in Neural Information Processing Systems, 33, 8174-8185.
\bibitem{ref6} Xu, Q., Sun, X., Wu, C. Y., Wang, P., \& Neumann, U. (2020). Grid-gcn for fast and scalable point cloud learning. In Proceedings of the IEEE/CVF Conference on Computer Vision and Pattern Recognition (pp. 5661-5670).
\bibitem{ref7} Y. Lyu, X. Huang and Z. Zhang, "Learning to Segment 3D Point Clouds in 2D Image Space," 2020 IEEE/CVF Conference on Computer Vision and Pattern Recognition (CVPR), 2020, pp. 12252-12261.
\bibitem{ref8} Li, M., \& Yin, D. (2017, May). A fast segmentation method of sparse point clouds. In 2017 29th Chinese Control And Decision Conference (CCDC) (pp. 3561-3565). IEEE.
\bibitem{ref9} Schumann, O., Hahn, M., Dickmann, J., \& Wöhler, C. (2018, July). Semantic segmentation on radar point clouds. In 2018 21st International Conference on Information Fusion (FUSION) (pp. 2179-2186). IEEE.
\bibitem{ref10} Jin, F., Sengupta, A., et al. (2020, April). Mmwave radar point cloud segmentation using gmm in multimodal traffic monitoring. In 2020 IEEE International Radar Conference (RADAR) (pp. 732-737). IEEE.
\bibitem{ref11} Adhikari, A., \& Sur, S. (2021, September). MilliPose: Facilitating Full Body Silhouette Imaging from Millimeter-Wave Device. In Adjunct Proceedings of the 2021 ACM International Joint Conference on Pervasive and Ubiquitous Computing and Proceedings of the 2021 ACM International Symposium on Wearable Computers (pp. 1-3).
\bibitem{ref12} Wang, Y., Sun, Y., Liu, Z., Sarma, S. E., Bronstein, M. M., \& Solomon, J. M. (2019). Dynamic graph cnn for learning on point clouds. Acm Transactions On Graphics (tog), 38(5), 1-12.
\bibitem{ref13} Li, Y., Bu, R., Sun, M., Wu, W., Di, X., \& Chen, B. (2018). PointCNN: Convolution On $\mathcal {X} $-Transformed Points. arXiv preprint arXiv:1801.07791.
\bibitem{ref14} Ma, C., Wojtowytsch, S., \& Wu, L. (2020). Towards a Mathematical Understanding of Neural Network-Based Machine Learning: what we know and what we don't. arXiv preprint arXiv:2009.10713.
\bibitem{ref15} Vaswani, A., Shazeer, N., Parmar, N., Uszkoreit, J., Jones, L., Gomez, A. N., ... \& Polosukhin, I. (2017). Attention is all you need. Advances in neural information processing systems, 30.
\bibitem{ref16} Zhao, H., Jiang, L., Jia, J., et al. (2021). Point transformer. In Proceedings of the IEEE/CVF International Conference on Computer Vision (pp. 16259-16268).
\bibitem{ref17} Lee, J., Lee, Y., Kim, J., Kosiorek, A., Choi, S.,  Teh, Y. W. (2019, May). Set transformer: A framework for attention-based permutation-invariant neural networks. In International conference on machine learning (pp. 3744-3753). 
\bibitem{ref18} Xue, H., Ju, Y., Miao, C., Wang, Y., Wang, S., Zhang, A., Su, L. (2021, June). mmMesh: Towards 3D real-time dynamic human mesh construction using millimeter-wave. In Proceedings of the 19th Annual International Conference on Mobile Systems, Applications, and Services (pp. 269-282).
\bibitem{ref19} James R. Munkres, Topology. Pearson, 2018.
\bibitem{ref20} Armeni, I., Sener, O., Zamir, A. R., Jiang, H., Brilakis, I., Fischer, M., \& Savarese, S. (2016). 3d semantic parsing of large-scale indoor spaces. In Proceedings of the IEEE conference on computer vision and pattern recognition (pp. 1534-1543).
\bibitem{ref21} Li, H., Liu, R., Wang, S., Jiang, W., \& Lu, C. X. (2022, September). Pedestrian Liveness Detection Based on mmWave Radar and Camera Fusion. In 2022 19th Annual IEEE International Conference on Sensing, Communication, and Networking (SECON) (pp. 262-270). IEEE.
\bibitem{ref22} Cao, D., Liu, R., Li, H., Wang, S., Jiang, W., \& Lu, C. X. (2022). Cross vision-rf gait re-identification with low-cost rgb-d cameras and mmwave radars. Proceedings of the ACM on Interactive, Mobile, Wearable and Ubiquitous Technologies, 6(3), 1-25.
\bibitem{ref23} Jiang, W., Li, F., Mei, L., Liu, R., \& Wang, S. (2022, September). VisBLE: Vision-Enhanced BLE Device Tracking. In 2022 19th Annual IEEE International Conference on Sensing, Communication, and Networking (SECON) (pp. 217-225). IEEE. 
\bibitem{yan20222} Yan H, Wang S, et al. $ o^{2} $-siterec: Store site recommendation under the o2o model via multi-graph attention networks. IEEE 38th International Conference on Data Engineering (ICDE). IEEE, 2022: 525-538. 
\bibitem{liu2022pred} Liu W, Ding Y, Wang S, et al. Para-Pred: Addressing Heterogeneity for City-Wide Indoor Status Estimation in On-Demand Delivery. Proceedings of the 28th ACM SIGKDD Conference on Knowledge Discovery and Data Mining. 2022: 3407-3417. 
\bibitem{liu2022exploiting} Liu Y, Guo B, Song X, et al. Exploiting Intra-and Inter-Region Relations for Sales Prediction via Graph Convolutional Network. In IEEE Global Communications Conference. IEEE, 2022: 3754-3759. 
\bibitem{wang-tkde-parking} Wang S, Zhu X, Wang G, et al. $\mathrm {W^{2}} $ Parking: A Data-Driven Win-Win Contract Parking Sharing Mechanism Under Both Supply and Demand Uncertainties[J]. IEEE Transactions on Knowledge and Data Engineering, 2022. 
\end{thebibliography}
\end{document}